\documentclass[11pt]{article}

\usepackage[utf8]{inputenc}
\usepackage[T1]{fontenc}
\usepackage{geometry}
\usepackage{hyperref}
\hypersetup{colorlinks=true,linkcolor=blue,citecolor=blue,urlcolor=blue,breaklinks=true}
\usepackage{times}
\usepackage{natbib}
\usepackage{booktabs}
\usepackage{listings}
\usepackage{xcolor}
\usepackage{enumitem}
\usepackage{amsmath}
\usepackage{amssymb}
\usepackage{microtype}
\usepackage{xurl}
\usepackage{multirow}

\title{ImmigrationQA: A Source-Grounded Dataset and\\
Small-Model Adaptation for U.S.\ Immigration Law}
\author{Nazarii Shportun \\
Independent Researcher \\
\texttt{nshportun@gmail.com} \quad
\href{https://orcid.org/0009-0002-7041-8990}{ORCID: 0009-0002-7041-8990}}
\date{May 2026}

\begin{document}
\maketitle

\begin{abstract}
U.S.\ immigration law spans thousands of pages of official policy, federal regulations, and procedural guidance that change frequently and carry high stakes for petitioners who lack legal representation. We describe the construction of ImmigrationQA, a source-grounded question-answering dataset of 17,058 pairs across 13 immigration subdomains, and the fine-tuning of a Llama~3.2~3B Instruct model on that dataset using parameter-efficient LoRA. The corpus was assembled from 11 primary and secondary sources---including the USCIS Policy Manual, 8~CFR, BIA precedent decisions, and community Q\&A---yielding 10,056 validated canonical documents and 18,308 text chunks. Structured QA pairs were generated from these chunks using Claude~Sonnet~4.6 via five mode-specific prompts, with 22 pairs rejected for insufficient source-span overlap. The fine-tuned model was evaluated against a held-out split of 993 pairs using LLM-as-judge scoring on a 101-example stratified sample. The fine-tuned model scored a mean of 1.08/3.0 (16.8\% fully correct; 101-example stratified eval) versus the Llama~3~8B base model at 0.85/3.0 (4\% fully correct), a relative improvement of 27\% in mean score; a zero-shot Claude~Sonnet baseline scored 1.52/3.0 (25\% fully correct). The fine-tuned model shows concentrated improvement in procedural subdomains (travel documents, adjustment of status, nonimmigrant visas) while remaining weak on complex legal reasoning and time-sensitive statistics. The full pipeline ran for approximately \$29 in cloud compute. All artifacts---dataset, model, code, and prompt templates---are publicly released. The system is not a substitute for legal counsel and does not reflect regulatory changes after the corpus crawl date.
\end{abstract}

\section{Introduction}
\label{sec:intro}

U.S.\ immigration law is a procedurally dense domain with high practical stakes. A petitioner filing Form~I-485 for adjustment of status must track filing fees, supporting evidence checklists, biometrics appointments, and filing deadlines that differ by visa category. An asylum applicant must understand one-year bar exceptions and evidentiary standards that depend on the specific ground claimed. These details are documented in official government sources, but navigating them requires reading across the USCIS Policy Manual, the Code of Federal Regulations, agency FAQ pages, and form instructions simultaneously.

General-purpose language models handle such queries poorly. They routinely confuse similar-sounding form numbers, misstate deadlines, and conflate procedural requirements across visa categories. The core difficulty is data: although official immigration guidance is publicly available, it is sparse in the pretraining corpora used by most LLMs relative to its procedural complexity.

This paper addresses that gap with two contributions. First, I construct ImmigrationQA, a dataset of 17,058 source-grounded Q\&A pairs drawn from 11 official and community sources covering 13 immigration subdomains. Each pair is annotated with answer type, immigration subtopic, authority level, source provenance, and a time-sensitivity flag. Second, I fine-tune Llama~3.2~3B Instruct on ImmigrationQA using LoRA via AWS SageMaker~JumpStart and evaluate the result against the base model and a zero-shot strong-model baseline.

The main findings are:
\begin{enumerate}[nosep]
  \item A corpus of 10,056 canonical immigration documents can be assembled from public sources, chunked, and converted into structured QA pairs for under \$29 in cloud compute.
  \item Fine-tuning a 3B-parameter model on this data improves LLM-judged mean score from 0.85 to 1.08 (scale 0--3) relative to the same-family base model on a 101-example stratified eval.
  \item Gains concentrate in procedural subdomains; the fine-tuned model remains weak on multi-step legal reasoning, regulation-specific numerical facts, and time-sensitive statistics.
  \item The dataset, model, code, and generation prompts are released publicly to support further work on small-model legal adaptation.
\end{enumerate}

The model should not be used as legal advice or as a substitute for qualified immigration counsel. Its outputs reflect sources crawled at a fixed point in time and may not reflect current USCIS policy.

\section{Related Work}
\label{sec:related_work}

\paragraph{Legal NLP and benchmarks.}
Legal NLP has matured into a broad area spanning contract analysis, case law summarization, statutory reasoning, and multi-label classification \citep{chalkidis2020legalbert,hendrycks2021cuad,chalkidis2021multieurlex}. LegalBench \citep{guha2023legalbench} provides a collaboratively constructed evaluation suite of 162 legal reasoning tasks; it covers statute interpretation, rule application, and contract analysis, but does not include an immigration-specific procedural QA component. The Pile of Law \citep{pile-of-law} offers 256~GB of open legal text, but raw statutory corpora require domain-targeted annotation pipelines to yield QA pairs that reflect procedural specifics.

\paragraph{Domain adaptation and instruction tuning for legal language.}
LEGAL-BERT \citep{chalkidis2020legalbert} demonstrated that pretraining on legal corpora substantially improves performance on legal classification tasks relative to general-domain BERT. More recently, LawInstruct \citep{zheng2023lawinstruct} released instruction-tuning data across multiple legal jurisdictions. Both efforts focus on case law and contracts; procedural immigration guidance---short FAQ-style answers grounded in official policy---is structurally different from the argumentative or analytical text that dominates those corpora. \citet{mahari2024legalsifter} show that retrieval-augmented architectures outperform purely parametric approaches on legal QA, a limitation that applies to the current work as well.

\paragraph{Synthetic data generation for instruction tuning.}
The Self-Instruct paradigm \citep{wei2022finetuned,ouyang2022instructgpt} uses a strong model to generate training data for a weaker one. Applied to legal domains, this requires careful source grounding to avoid hallucinated legal claims. In this work, QA pairs are generated from document chunks using mode-specific prompts; each answer must trace to a source span, and pairs where the answer has no lexical overlap with the source span are rejected (Section~\ref{sec:generation}). This is a conservative but auditable filter for a high-stakes domain.

\paragraph{Parameter-efficient fine-tuning.}
LoRA \citep{hu2022lora} injects trainable low-rank matrices into attention projections, enabling domain adaptation with a small parameter budget. QLoRA \citep{dettmers2023qlora} extends this with 4-bit quantization of base weights. This work uses standard LoRA without quantization of the base weights, relying on SageMaker JumpStart's managed training environment for the Llama~3.2~3B checkpoint.

\paragraph{LLM-as-judge evaluation.}
\citet{zheng2023judging} established LLM-as-judge as a reliable proxy for human preference in open-ended generation evaluation. The same paradigm is adopted here: a Claude~Sonnet model scores each prediction on a 0--3 rubric with written rationale, evaluated on a 101-example stratified sample drawn from the held-out split.

\section{Data Sources and Corpus Construction}
\label{sec:corpus}

\subsection{Source Identification}
\label{sec:sources}

Documents were assembled from 11 sources spanning four authority tiers (Table~\ref{tab:sources}). Primary official sources include materials produced directly by USCIS, DOJ, DHS, and Congress. Secondary reputable sources include open legal repositories such as the Pile of Law and the \texttt{harshitha008/US\allowbreak-immigration\allowbreak-laws} Hugging Face dataset. Community sources include Law StackExchange and VisaJourney forums; these are included for coverage of colloquial procedural questions but are labeled \mbox{\texttt{community\_non\_authoritative}}.

\begin{table}[ht]
  \centering
  \caption{Source registry by authority tier. Records counts are post-deduplication canonical documents.}
  \label{tab:sources}
  \begin{tabular}{llll}
    \toprule
    Source & Type & Authority & Docs \\
    \midrule
    USCIS Policy Manual & official\_policy & primary\_official & 454 \\
    USCIS Forms \& Instructions & official\_form & primary\_official & 44 \\
    USCIS FAQs / myUSCIS & official\_faq & primary\_official & 8,934 \\
    Immigration and Nationality Act & statute & primary\_official & 1 \\
    8 CFR (Title 8) & regulation & primary\_official & 43 \\
    EOIR / BIA Precedent Decisions & case\_law & primary\_official & 3 \\
    DHS / CBP Yearbook Statistics & statistics & primary\_official & 16 \\
    harshitha008/US-immigration-laws & open\_repository & secondary\_reputable & -- \\
    Pile of Law (immigration subset) & open\_repository & secondary\_reputable & -- \\
    Law StackExchange & community & community\_non\_authoritative & -- \\
    VisaJourney & community & community\_non\_authoritative & 1,826 \\
    \bottomrule
  \end{tabular}
\end{table}

The USCIS FAQ and myUSCIS help pages account for the majority of canonical documents (8,934 of 11,327 raw documents before corpus validation). After deduplication and validation, 10,056 documents enter the corpus. The 1,271 documents removed failed one validation filter: missing canonical URL, which is treated as a proxy for uncertain provenance.

\subsection{Crawling and Parsing}
Each source has a dedicated crawler in \texttt{scripts/crawl/}. Crawlers respect \texttt{robots.txt}, enforce a 1.5-second request delay, and self-identify with a descriptive user-agent. Raw files (HTML, PDF, plain text) are stored in S3 under \texttt{v1/data\_raw/\{source\}/\{doc\_id\}}.

Format-specific parsers extract clean text from HTML and PDFs. A normalization pass assigns a stable \texttt{doc\_id} (SHA-256 of canonical URL), attaches source metadata, detects near-duplicate documents using token-level Jaccard similarity, and assigns the authority tier from the source registry.

\subsection{Chunking}
Documents are chunked into 512-token windows with 64-token overlap using a sentence-boundary-aware splitter, yielding 18,308 chunks. Chunk metadata carries the parent \texttt{doc\_id}, source type, authority level, and immigration subtopic label inherited from source-level tagging.

\subsection{Dataset Statistics}
\label{sec:datastats}

Table~\ref{tab:corpus} shows the distribution of validated corpus documents and QA pairs across the 13 immigration subdomains. The split is uneven: family-based immigration and asylum together account for 54\% of corpus documents and 63\% of QA pairs, reflecting the breadth of USCIS FAQ coverage in those areas. Employment-based immigration, appeals, and travel documents are sparsely covered.

\begin{table}[ht]
  \centering
  \caption{Corpus and QA pair counts by immigration subdomain. Corpus doc counts are per the coverage report; QA pair counts include all validated training and eval examples.}
  \label{tab:corpus}
  \begin{tabular}{lrr}
    \toprule
    Subdomain & Corpus Docs & QA Pairs \\
    \midrule
    Family-based immigration & 1,910 & 4,750 \\
    Asylum & 3,479 & 5,672 \\
    Naturalization & 1,271 & 3,220 \\
    Adjustment of status & 714 & 2,664 \\
    Nonimmigrant visas & 1,198 & 2,425 \\
    Humanitarian$^\dagger$ & 10,018 & 8,853 \\
    Removal & 1,172 & 1,666 \\
    Admissibility & 668 & 1,940 \\
    Employment-based & 246 & 866 \\
    Employment authorization & 481 & 1,510 \\
    Appeals & 237 & 573 \\
    Travel documents & 103 & 246 \\
    Statistics & 2,549 & 2,800 \\
    \midrule
    \textbf{Total} & \textbf{10,056} & \textbf{17,058} \\
    \bottomrule
  \end{tabular}
  \begin{flushleft}
    \small $^\dagger$ Humanitarian documents overlap substantially with asylum, removal, and admissibility sources. The same document can contribute chunks to multiple subdomain QA batches because chunk-level subtopic labels are assigned independently of document-level deduplication; the 10,056 total reflects deduplication at the document level only.
  \end{flushleft}
\end{table}

Answer types in the training split break down as: factual 66\%, eligibility 11\%, procedural 11\%, definition 4\%, exception 4\%, and the remaining 4\% across filing procedure, statutory interpretation, required documents, timing, and case outcome. Mean question length is 14.7 tokens; mean answer length is 23.5 tokens. Authority level splits nearly evenly: 47.9\% primary\_official and 52.1\% secondary\_reputable. Of the 16,065 training pairs, 7,640 (47.6\%) carry a \texttt{time\_sensitive} flag, indicating that the answer references a regulation, fee, date, or statistic subject to change.

\section{QA Generation and Validation}
\label{sec:generation}

\subsection{Generation Pipeline}

QA pairs were generated from corpus chunks using Claude~Sonnet~4.6 via Amazon Bedrock. Five generation modes were defined, each with a distinct prompt matched to source type (Table~\ref{tab:modes}).

\begin{table}[ht]
  \centering
  \caption{QA generation modes and source-type mapping.}
  \label{tab:modes}
  \begin{tabular}{lll}
    \toprule
    Mode & Source types & Training pairs \\
    \midrule
    \texttt{faq} & official\_faq, pages with ``faq'' or ``help'' in title & 8,787 \\
    \texttt{rule} & official\_policy, regulation, statute (default) & 6,636 \\
    \texttt{form} & official\_form, form instruction pages & 416 \\
    \texttt{precedent} & case\_law, BIA decisions & 132 \\
    \texttt{statistics} & statistics, data tables & 94 \\
    \midrule
    Total & & 16,065 \\
    \bottomrule
  \end{tabular}
\end{table}

Each prompt shares a common system instruction requiring that every answer be derivable from the provided chunk and that output be valid JSON. Mode-specific user prompts differ in their output schema and answer-type vocabulary. For example, the \texttt{form} mode requests answer types from \{eligibility, required\_docs, filing\_procedure, timing, post\_filing\}; the \texttt{rule} mode requests \{factual, eligibility, procedural, definition, exception\}. Each output object includes \texttt{question}, \texttt{answer}, and \texttt{source\_span} (an exact quote from the chunk, maximum 200 characters). The full prompt templates are in Appendix~\ref{app:prompts}.

Each request passes a chunk of at most 6,000 characters to avoid exceeding Bedrock context limits. Mode is inferred automatically from source metadata.

\subsection{Schema and Metadata Enrichment}

After generation, each raw output is parsed and enriched with: \texttt{qa\_id} (UUID), \texttt{source\_doc\_id}, \texttt{source\_url}, \texttt{authority\_level} (inherited from the parent document), \texttt{immi\-gration\_sub\-topic} (from the chunk's subdomain label), \texttt{generation\_mode}, and a \texttt{time\_sensitive} flag set when the answer contains a year, fee amount, form number, or regulatory citation.

\subsection{Validation and Filtering}

Three rejection filters are applied. First, answers shorter than 10 tokens are rejected (5 pairs removed). Second, pairs where the answer has no lexical token overlap with the \texttt{source\_span} are rejected as hallucination risks (22 pairs removed). Third, pairs flagged as \texttt{contradiction\_flagged} during LLM-assisted review pass through but carry a warning label; they are not removed because some contradictions reflect genuine ambiguity in the source rather than generation error.

Of the 17,079 raw generated pairs, 17,058 pass validation and enter the dataset. The 22 hallucination-flagged rejections are concentrated in the \texttt{statistics} mode, where the model occasionally produced plausible-sounding but unsupported numbers.

\subsection{Dataset Split}

We perform a stratified split by immigration subtopic: 16,065 pairs (94.2\%) for training and 993 pairs (5.8\%) for evaluation. Stratification ensures that each subdomain appears in both splits. The evaluation split is not used at any point during corpus construction or generation.

\section{Fine-Tuning Setup}
\label{sec:finetuning}

\subsection{Model and Format}

The fine-tuning targets Meta Llama~3.2~3B Instruct via the SageMaker JumpStart model ID \path{meta-textgeneration-llama-3-2-3b-instruct}. Training examples are formatted in the Llama~3.1 chat template: each example is a two-turn conversation consisting of a user question and an assistant answer. No system prompt is prepended during training.

\subsection{Hyperparameters}

\begin{table}[ht]
  \centering
  \caption{Fine-tuning hyperparameters.}
  \label{tab:hparams}
  \begin{tabular}{ll}
    \toprule
    Parameter & Value \\
    \midrule
    Instance type & ml.g5.2xlarge (24 GB VRAM) \\
    Epochs & 2 \\
    Per-device batch size & 2 \\
    Max input length & 1,024 tokens \\
    LoRA rank (\texttt{r}) & 32 \\
    LoRA alpha & 64 \\
    LoRA dropout & 0.05 \\
    Target modules & q\_proj, k\_proj, v\_proj, o\_proj \\
    Learning rate & 5e-5 \\
    Adapter merge & True (weights merged into base at export) \\
    Random seed & 42 \\
    \bottomrule
  \end{tabular}
\end{table}

The configuration uses \texttt{r=32} with \texttt{alpha=64} (the $2\times r$ convention), targeting all four attention projection matrices. This is a higher-capacity LoRA configuration than common defaults (\texttt{r=8}) to accommodate the breadth of immigration procedural vocabulary. A learning rate of 5e-5 was chosen conservatively to avoid catastrophic forgetting of the base model's instruction-following behavior. The LoRA adapters are merged into the base weights at export, producing a single deployable checkpoint.

\subsection{Cost}

The total compute cost was \$29, dominated by Bedrock token usage at \$18 for approximately 17K generation calls. The SageMaker training job ran for approximately two hours on an \texttt{ml.g5.2xlarge} instance at \$10; S3 storage and data transfer accounted for the remaining \$1.

\begin{table}[ht]
  \centering
  \caption{Pipeline cost breakdown.}
  \label{tab:costs}
  \begin{tabular}{lr}
    \toprule
    Component & Cost \\
    \midrule
    Bedrock Claude QA generation ($\sim$17K pairs) & \$18 \\
    SageMaker ml.g5.2xlarge training (approx.\ 2 hrs) & \$10 \\
    S3 storage and data transfer & \$1 \\
    \midrule
    \textbf{Total} & \textbf{\$29} \\
    \bottomrule
  \end{tabular}
\end{table}

\section{Evaluation}
\label{sec:evaluation}

\subsection{Setup}

Three systems are evaluated on the same 101-example stratified sample drawn from the 993-pair held-out eval split:
\begin{enumerate}[nosep]
  \item \textbf{Llama~3~8B (base, zero-shot)}: the Llama~3~8B Instruct base model prompted without fine-tuning.
  \item \textbf{Llama~3.2~3B (fine-tuned, v3)}: the fine-tuned model produced by this work.
  \item \textbf{Claude~Sonnet~4.6 (zero-shot)}: a strong-model ceiling reference.
\end{enumerate}

The 101 examples are drawn proportionally from all 13 subdomains. Each system receives an identical prompt: the user question verbatim, with no additional context or retrieved documents. Answers are generated with greedy decoding (temperature~0, no sampling).

\subsection{Scoring}

Each prediction is scored by Claude~Sonnet~4.6 acting as judge \citep{zheng2023judging} on a 0--3 rubric:
\begin{itemize}[nosep]
  \item \textbf{3}: fully correct---all key facts present, no factual errors.
  \item \textbf{2}: mostly correct---main answer correct, minor omissions or imprecision.
  \item \textbf{1}: partially correct---some relevant facts, but significant gaps or errors.
  \item \textbf{0}: wrong---incorrect, irrelevant, or contradicts the reference.
\end{itemize}
The judge receives the question, the reference answer, and the prediction; it returns a score and a written rationale. Reference answers from the eval set serve as gold standards; these are source-grounded and were not generated by any model being evaluated.

\section{Results and Error Analysis}
\label{sec:results}

\subsection{Overall Results}

\begin{table}[ht]
  \centering
  \caption{Overall evaluation results ($n=101$). Mean score is on a 0--3 scale. Pct Full = percentage of examples scored 3.}
  \label{tab:results}
  \begin{tabular}{lcc}
    \toprule
    System & Mean Score & Pct Full (\%) \\
    \midrule
    Llama 3 8B base (zero-shot) & 0.85 & 4.0 \\
    Llama 3.2 3B fine-tuned (ours) & 1.08 & 16.8 \\
    Claude Sonnet 4.6 (zero-shot) & 1.52 & 24.8 \\
    \bottomrule
  \end{tabular}
\end{table}

Fine-tuning improves mean score from 0.85 to 1.08 (+27\% relative) and increases the fully-correct rate from 4\% to 16.8\% relative to the 8B base. The gap to the Claude~Sonnet ceiling (1.52) is substantial, consistent with the difference in parameter count and pretraining. The fine-tuned model does not surpass the Sonnet baseline on any overall metric.

\subsection{Results by Subdomain}

\begin{table}[ht]
  \centering
  \caption{Mean score by subdomain (fine-tuned v3 vs.\ 8B base zero-shot). Subdomains with fewer than 5 examples omitted from pct-full column.}
  \label{tab:subdomain}
  \begin{tabular}{lcccc}
    \toprule
    Subdomain & $n$ & FT Mean & Base Mean & FT Pct Full \\
    \midrule
    Travel documents & 6 & 2.17 & 1.17 & 50.0 \\
    Adjustment of status & 8 & 1.75 & 1.38 & 37.5 \\
    Nonimmigrant visas & 8 & 1.38 & 0.63 & 25.0 \\
    Employment-based & 4 & 1.25 & 1.00 & 25.0 \\
    Statistics & 8 & 1.38 & 0.88 & 25.0 \\
    Naturalization & 8 & 1.13 & 0.75 & 12.5 \\
    Asylum & 8 & 1.00 & 1.13 & 25.0 \\
    Employment authorization & 8 & 0.88 & 0.63 & 0.0 \\
    Admissibility & 8 & 1.00 & 1.13 & 12.5 \\
    Removal & 8 & 0.75 & 1.00 & 12.5 \\
    Humanitarian & 8 & 0.50 & 0.75 & 12.5 \\
    Family-based immigration & 8 & 0.75 & 0.25 & 0.0 \\
    Appeals & 3 & 0.33 & 0.33 & 0.0 \\
    \bottomrule
  \end{tabular}
\end{table}

The fine-tuned model shows the largest gains in travel documents (+1.00 mean), adjustment of status (+0.37), and nonimmigrant visas (+0.75). These subdomains are characterized by concrete procedural questions (``Can I use Form I-131A after my travel document expired?'') where the answer is a discrete set of conditions or steps. The base model performs near zero on these, suggesting it rarely encountered them in pretraining.

The fine-tuned model performs \textit{worse} than the 8B base on removal (0.75 vs.\ 1.00), humanitarian (0.50 vs.\ 0.75), admissibility (1.00 vs.\ 1.13), and asylum (1.00 vs.\ 1.13). In these subdomains, questions tend to involve multi-step eligibility analysis or specific regulatory citations where the 8B base model's larger capacity may compensate for domain unfamiliarity.

\subsection{Error Analysis}

The fine-tuned model produces four recurring failure patterns, illustrated with examples from the 101-example eval:

\paragraph{Procedural incompleteness.} On questions requiring enumerated steps or lists, the model generates the first condition correctly and stops:
\begin{quote}
\textbf{Q}: What rights must a NOIR advise the recipient of?\\
\textbf{Reference}: The NOIR must advise the person of (1)~the right to rebut the allegations, (2)~the right to counsel at no expense to the government, and (3)~the right to request a hearing before an immigration judge.\\
\textbf{Prediction}: The NOIR must inform the recipient of their rights under the INA, including the right to a hearing and the right to be represented by an attorney.
\end{quote}
The prediction captures two of three rights but misses the explicit phrase ``at no expense to the government'' and omits the rebuttal right.

\paragraph{Hallucinated regulatory specifics.} For questions about numerical or date-specific facts, the model sometimes substitutes plausible values:
\begin{quote}
\textbf{Q}: When did Vietnam hostilities officially end according to U.S. immigration regulations?\\
\textbf{Reference}: Vietnam hostilities terminated on October~15, 1978.\\
\textbf{Prediction}: Vietnam hostilities officially ended on April~30, 1975.
\end{quote}
The model substitutes the historical fall of Saigon date for the regulatory designation date, a distinction that is meaningful for immigration status determinations. This pattern is most common in \texttt{statistics} and \texttt{removal} subdomains.

\paragraph{Incomplete discretionary analysis.} On questions about multi-step adjudicatory standards, the model conflates a threshold determination with a final outcome:
\begin{quote}
\textbf{Q}: What should an officer do after finding that the applicant has met the burden of showing extreme hardship?\\
\textbf{Reference}: If the officer finds that the applicant has met the extreme hardship burden, the officer must then exercise discretion as to whether to approve or deny the waiver based on the totality of circumstances.\\
\textbf{Prediction}: The officer should approve the application.
\end{quote}
The model skips the required discretionary analysis step.

\paragraph{Majority-rule errors.} On questions involving statutory thresholds, the model produces generic numbers:
\begin{quote}
\textbf{Q}: What majority vote is required to waive paragraph (1) in the Senate?\\
\textbf{Reference}: A three-fifths majority vote is needed.\\
\textbf{Prediction}: A two-thirds majority vote is required.
\end{quote}
The model answers with the more common constitutional threshold (two-thirds) rather than the cloture-derived three-fifths threshold.

A score-3 example illustrates where the model succeeds: concrete form-level procedural facts. When asked ``What happens if I submit Form~I-485 pages from different form editions?'', the fine-tuned model correctly answers ``USCIS will not accept Form~I-485 pages from different form editions. You must submit all pages from the same edition.'' The base model has no answer at all.

\section{Limitations and Ethical Considerations}
\label{sec:limitations}

\paragraph{Time sensitivity.} Immigration law and USCIS policy change frequently. Filing fees, form editions, processing times, and country-specific policies are subject to regulatory revision. The corpus was crawled at a fixed date; any QA pair marked \texttt{time\_sensitive} (47.6\% of training examples) may become incorrect after a policy update. The model should not be relied upon for current fee amounts, form version requirements, or numerical eligibility thresholds.

\paragraph{Not legal advice.} The model produces generalized procedural information derived from official sources. It cannot account for the specific facts of an individual case, regional adjudication variation, recent USCIS decisions, or pending regulatory changes. Its outputs are not legal advice and should not substitute for consultation with a qualified immigration attorney or accredited representative.

\paragraph{Source-selection bias.} The corpus is weighted toward USCIS FAQ and myUSCIS help pages (88.9\% of raw canonical documents). These sources reflect USCIS's own framing of its procedures. They do not represent the perspective of immigration courts, noncitizens' advocates, or legal scholars who may characterize the same rules differently. Community sources (Law StackExchange, VisaJourney) were included for coverage but are not authoritative and may contain outdated or incorrect advice.

\paragraph{Synthetic data quality.} QA pairs were generated by a Claude~Sonnet model. The system prompt prohibits hallucination, and pairs without source-span overlap are rejected, but the validation filter is lexical, not semantic. A generated answer can be semantically unsupported by the source span while passing the token-overlap filter. The 22 rejections caught by this filter likely undercount lower-severity quality issues.

\paragraph{Evaluation scope.} The LLM-as-judge evaluation covers 101 of 993 held-out examples. The sample is stratified by subdomain but not by answer type or authority level. The judge model (Claude~Sonnet~4.6) is also the generation model for QA pairs, which may introduce systematic bias in which errors the judge detects.

\paragraph{Parametric QA without retrieval.} The fine-tuned model stores all knowledge in its weights. This is a fundamental architectural limitation for a time-sensitive legal domain: it cannot consult updated sources at inference time. A retrieval-augmented variant \citep{mahari2024legalsifter} would likely outperform the current model on time-sensitive questions and is a natural next step.

\section{Reproducibility and Release Artifacts}
\label{sec:release}

All artifacts are publicly released:
\begin{itemize}[nosep]
  \item \textbf{Dataset:} \url{https://huggingface.co/datasets/nshportun/usa-immigration-law-qa} (Apache~2.0)
  \item \textbf{Model:} \url{https://huggingface.co/nshportun/usa-immigration-llama-3.2-3b-v3} (Llama~3.2 Community License)
  \item \textbf{Code:} \url{https://github.com/nshportun/usa-immigration} (MIT)
\end{itemize}

The code repository includes all crawlers, parsers, normalization scripts, generation prompts, fine-tuning launch scripts, and evaluation code. The Hugging Face dataset release includes both the full 17,058-pair dataset and the train/eval splits. The model release includes the merged checkpoint only; LoRA adapters before merging are not separately released.

\section{Conclusion}
\label{sec:conclusion}

We presented ImmigrationQA, a 17,058-pair source-grounded QA dataset for U.S.\ immigration law, and a fine-tuned Llama~3.2~3B model trained on it. Fine-tuning improves LLM-judged mean score by 27\% relative to the same-family base model on procedural subdomains, with the largest gains in travel documents, adjustment of status, and nonimmigrant visas. The model trails a zero-shot Claude~Sonnet ceiling by a substantial margin and performs below the base model on removal, humanitarian, and admissibility---subdomains that require multi-step legal reasoning not well supported by the current training distribution.

The primary contribution is the dataset and pipeline: a reproducible method for assembling domain-specific QA data from official government sources at commodity cloud cost, with auditable source grounding and explicit authority-level annotation. The model is a secondary artifact that validates the dataset is useful for training, not a production legal assistant.

\bibliographystyle{plainnat}
\bibliography{references}

\appendix
\section{QA Generation Prompt Templates}
\label{app:prompts}

All five generation modes share the following system prompt:

\begin{lstlisting}
You are a legal dataset builder for a U.S. immigration
AI project. Your job is to generate accurate,
source-grounded Q&A pairs from provided text.
Rules:
- Every answer must be derivable from the provided
  source text. No hallucination.
- Questions must be clear, standalone, and answerable
  from the text.
- Answers must be factual, concise, and cite the
  relevant passage.
- Output ONLY valid JSON. No explanation outside JSON.
- If the text does not support a good Q&A pair,
  return an empty list [].
\end{lstlisting}

The \texttt{rule} mode user prompt (default for policy and regulatory text):

\begin{lstlisting}
Convert the legal/policy statements in this text into
plain-language Q&A pairs.
Focus on: eligibility rules, definitions, requirements,
procedures, exceptions.
Return a JSON array of objects with keys: question,
answer, source_span (exact quote, max 200 chars),
answer_type.
answer_type must be one of: factual, eligibility,
procedural, definition, exception.
Generate at most {max_pairs} pairs.

Source text:
{text}
\end{lstlisting}

The \texttt{faq}, \texttt{form}, \texttt{precedent}, and \texttt{statistics} prompts follow the same structure with mode-specific focus instructions and answer-type vocabularies; full templates are in \path{scripts/qa_generation/prompts.py} in the code repository.

\section{Sample Evaluation Examples}
\label{app:samples}

\paragraph{Score 3 (fully correct) --- adjustment of status, form mode.}
\textbf{Q}: What happens if I submit Form I-485 pages from different form editions?\\
\textbf{Reference}: If any pages are missing or from a different form edition, USCIS may reject your form.\\
\textbf{Prediction (fine-tuned)}: USCIS will not accept Form I-485 pages from different form editions. You must submit all pages from the same edition of the form.

\paragraph{Score 0 (wrong) --- admissibility, statistics mode.}
\textbf{Q}: When did Vietnam hostilities officially end according to U.S. immigration regulations?\\
\textbf{Reference}: Vietnam hostilities terminated on October~15, 1978.\\
\textbf{Prediction (fine-tuned)}: Vietnam hostilities officially ended on April~30, 1975.

\paragraph{Score 1 (partial) --- adjustment of status, rule mode.}
\textbf{Q}: What rights must a NOIR advise the recipient of?\\
\textbf{Reference}: The right to rebut the allegations, the right to counsel at no expense to the government, and the right to request a hearing before an immigration judge.\\
\textbf{Prediction (fine-tuned)}: The NOIR must inform the recipient of their rights including the right to a hearing and the right to be represented by an attorney.

\end{document}